\documentclass[conference]{IEEEtran}
\IEEEoverridecommandlockouts
\usepackage[hyphens]{url}  % DO NOT CHANGE THIS
\usepackage{graphicx} % DO NOT CHANGE THIS
\usepackage{caption} % DO NOT CHANGE THIS AND DO NOT ADD ANY OPTIONS TO IT
\usepackage{todonotes} %Zlatan
\usepackage[linesnumbered,ruled,vlined]{algorithm2e}
\usetikzlibrary{calc,quotes,angles}
\usetikzlibrary{calc}
\usetikzlibrary{decorations.markings}
\usepackage{amsmath}

\def\BibTeX{{\rm B\kern-.05em{\sc i\kern-.025em b}\kern-.08em
    T\kern-.1667em\lower.7ex\hbox{E}\kern-.125emX}}
\usepackage{subcaption}

\title{\LARGE \bf A Multi-Heuristic Search-based Motion Planning \\ for Automated Parking
\thanks{*Work done while at Virtual Vehicle Research GmbH.}
}

\author{\IEEEauthorblockN{1\textsuperscript{st} Bhargav Adabala}
\IEEEauthorblockA{\textit{Virtual Vehicle Research GmbH} \\
Graz, Austria\\
bhargav.adabala@gmail.com}
\and
\IEEEauthorblockN{2\textsuperscript{nd} Zlatan Ajanovi\'{c}}
\IEEEauthorblockA{\textit{Virtual Vehicle Research GmbH} \\
Graz, Austria \\
0000-0003-2214-0007}
}

\begin{document}

\maketitle
\thispagestyle{empty}
\pagestyle{empty}

%Finished
\begin{abstract}
% Planning is a crucial component of autonomous vehicle control. It is responsible for finding a collision-free sequence of states that take the vehicle to its goal. 
In unstructured environments like parking lots or construction sites, due to the large search-space and kinodynamic constraints of the vehicle, it is challenging to achieve real-time planning.
Several state-of-the-art planners utilize heuristic search-based algorithms. However, they heavily rely on the quality of the single heuristic function, used to guide the search. Therefore, they are not capable to achieve reasonable computational performance, resulting in unnecessary delays in the response of the vehicle. 
In this work, we are adopting a Multi-Heuristic Search approach, that enables the use of multiple heuristic functions and their individual advantages to capture different complexities of a given search space. Based on our knowledge, this approach was not used previously for this problem. For this purpose, multiple admissible and non-admissible heuristic functions are defined, the original Multi-Heuristic A* Search was extended for bidirectional use and dealing with hybrid continuous-discrete search space, and a mechanism for adapting scale of motion primitives is introduced. 
To demonstrate the advantage, the Multi-Heuristic A* algorithm is benchmarked against a very popular heuristic search-based algorithm, Hybrid A*. The Multi-Heuristic A* algorithm outperformed baseline in both terms, computation efficiency and motion plan (path) quality.
\end{abstract}

\begin{IEEEkeywords}
Motion Planning, Automated Driving, Multi-Heuristic Search, A* Search
\end{IEEEkeywords}

\section{Introduction}
%\todo[inline]{rewrite smoothly}

Robot motion planning problems can be elegantly formulated as path planning in higher-dimensional configuration space \cite{lozanoperez1979algorithm}. However, finding a solution is computationally challenging due to a large continuous search space and kinodynamic constraints. The sampling-based approaches for motion planning have been extensively studied in robotics \cite{kingston2018sampling}. Instead of explicitly constructing the collision-free configuration space, which is time-consuming to compute, these algorithms probe the free space and search with a sampling strategy. The algorithms stop when a path connecting the initial and final poses is found. According to the sampling type, the sampling-based path-finding algorithms can be classified into two categories: random-sampling-based algorithms and orderly-sampling-based algorithms.

The most popular random-sampling-based algorithm is the Rapidly-exploring Random Tree (RRT) \cite{lavalle1998rapidly}. RRT can be considered as a special case of Monte Carlo Tree Search (MCTS) \cite{browne2012survey}.
The most notable orderly sampling-based algorithm is A* \cite{hart1968formal}, including many of its extensions. It was initially developed to plan a path for the Shakey robot and it was further generalized and used for many different domains since then. 

%\begin{figure}
%	\includegraphics[width=\columnwidth]{figures/single_track_model.png}
%	\centering
%	\caption{\small Motion Planning for Autonomous Parking.}
%		\todo[inline]{Redraw}
%	\label{fig:eyecatcher}
%\end{figure}

%-----------------------------------------------------------------
\begin{figure}
  \centering
	\includegraphics[width= \columnwidth]{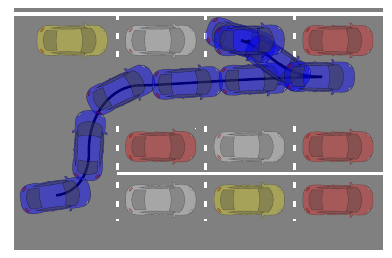}
  \caption{\small Motion Planning for Autonomous Parking.}
  \label{fig:eyecatcher}
          
\end{figure}
%-----------------------------------------------------------------

%ICAPS? \cite{ferguson2005guide} - survey like

The orderly sampling-based algorithms tend to be more efficient than random sampling-based planners, especially when the dimensions of the state spaces are fewer than six \cite{lavalle2004on}. Random sampling-based planners inherently come with the disadvantages of being highly non-deterministic and converging towards solutions that are far from the optimum and suffer from bug trap problems \cite{karaman2010incremental}. The variants of RRT such as RRT* also provide comparable solutions, but they are computationally expensive and the efficiency depends on the size of the search space \cite{karaman2011anytime}. Furthermore, if no collision-free path to the goal exists, orderly sampling-based algorithms can report this failure much more quickly than random sampling-based ones. However, original A* deals with discrete state-space and requires discretization of the configuration space. Continuous state-space and kinodynamic constraints can be satisfied by constructing a lattice in the form of a regular grid \cite{pivtoraiko2005generating}. %update
Another approach is to use motion primitives \cite{frazzoli2002real}. To avoid the problem of rounding states generated using motion primitives to the grid, Hybrid-State A* \cite{dolgov2008practical} might be used.

%Car

%Dubins curve
%\cite{dubins1957on}
%
%\cite{reeds1990optimal}

%%classics, before
%\cite{scheuer1996continuous}
%\cite{scheuer1998planning}

%Classic darpa
Sampling-based motion planning approaches were extensively used for autonomous vehicle path planning in unstructured environments during the 2007 DARPA Urban Challenge, both A*-based \cite{likhachev2009planning}, \cite{dolgov2008practical} and RRT-based \cite{kuwata2009real}.
%Recent AD parking
Several extensions for these approaches have been introduced in recent years, namely A*-based \cite{nemec2019improving, dai2021longhorizona, leu2022autonomousa},
RRT*-based \cite{banzhaf2017hybrid} 
and optimization-based \cite{zhang2020optimization}.
An overview of recent developments and open challenges is presented in \cite{banzhaf2017future}.

% recent search-based
Besides motion planning for wheeled vehicles in unstructured environments (e.g. autonomous parking), different variants of search-based planning were recently used for planning footsteps for humanoid robots \cite{ranganeni2020effective},
%MHA*
robot manipulation \cite{mandalika2018lazy}, underwater vehicles \cite{youakim2020multirepresentation}, 
%MHA*
the aggressive flight of UAVs \cite{liu2018search},
as well as for special use-cases in automated driving such as energy-efficient driving \cite{ajanovic2018novel}, driving in complex scenarios in structured urban environments \cite{ajanovic2018search}, unstructured and partially observable environments \cite{dai2021longhorizona}
and performance driving including drifting maneuvers \cite{ajanovic2019search}\cite{ajanovic2023search}.
%, where motion primitives are generated based on equilibrium state manifold as extension of closed-loop (\cite{kuwata2009real}).

%A* connect only 9 citations \cite{islam2016Astarconnect}

%\subsection{Contribution}
%There is no single best approach to solve the motion planning problem of autonomous parking. Even though, in research geometric approaches are more popular, it involves driver actions and cannot be used as fully automated solution. But, they provide a very good detail of physical constraints of the vehicle under control. Similarly, random sampling approaches can provide good quality solutions, but they need to be adapted or modified to properly deal with the depression regions i.e., regions in the search space where the algorithm does not correlate well with the actual cost-to-goal values. The orderly sampling-based approaches like A* provides a better solution for path planning problem depending on the heuristic function. But, it is not practical to define a single heuristic function that can capture all complexities of the search space.
This paper presents a  motion planning approach for wheeled vehicles in unstructured environments based on Multi-Heuristic Search \cite{aine2016multi}. %cite
\textit{Based on our knowledge, this is the first application of a Multi-Heuristic Search for motion planning in automated parking scenarios.} %\par
This is achieved by using a combination of geometric and orderly sampling-based approaches in order to achieve maximum coverage of the complexities within the search space.  
For this purpose, two heuristic functions are defined to get an accurate estimate of the cost-to-go and prune the unnecessary search nodes for faster path computation. The geometric approach is used to solve the simplified problem (without obstacles) by modeling the physical constraints of the vehicle and is used as a second heuristic function in a Multi-Heuristic A* algorithm, the first being the path length to the destination while considering obstacles but neglecting some physical constraints like turning radius. With this approach, both non-holonomic and holonomic constraints are combined to provide an optimal solution to the motion planning problem.
Two approaches for the solution have been developed, one using \texttt{Forward Search} and the other using \texttt{Bi-directional Search}. 
Additionally, adaptive motion primitive arc length was developed to avoid the search getting stuck in depression regions indefinitely.
An earlier version of this work is presented in the ICAPS 2020 PlanRob workshop \cite{adabala2020multi}.

\section{Autonomous Parking as Motion Planning Problem}
\label{ch:Problem_Formulation}

%\todo[inline]{rewrite smoothly}

%Typically, motion planning for autonomous parking involves two steps \textbf{\textit{Vehicle Guidance}} and \textbf{\textit{Parking Manoeuvre}}. In the Vehicle Guidance step, the vehicle shall be guided to the desired parking slot from a given starting point within or outside the parking lot. Once the vehicle approaches the desired parking slot, the vehicle shall then be maneuvered into the parking slot. In order to perform the above steps, it is important to understand the environment of the parking lot. This can be done \textbf{\textit{Onboard}} or \textit{\textbf{Offboard}}. 

%In the \textbf{\textit{Onboard}} approach, the driver drives the vehicle through the parking lot, and on-board sensors scan and detect an empty parking slot as the driver passes the empty slot. After the detection of the empty slot, the vehicle control unit generates the required steering angles automatically and suggests the driver, to actuate the pedals and gear lever position. Almost all the commercially available solutions use this approach e.g BMW ADAS Parking Assistant \cite{wahl2008developing}. It can be seen that with this approach the autonomy is limited and the driver should handle the Vehicle Guidance manually.

The autonomous parking problem tackled in this work represents a fully observable problem where intelligent infrastructure provides a connected vehicle with information about the structure and area of the parking lot. The driver enters the parking and drives the vehicle into a designated drop-off area. The autonomous parking algorithm overtakes a control and guides and maneuvers the vehicle into the assigned parking slot automatically based on the information from the infrastructure. This concept was demonstrated by Bosch and Daimler as Automated Valet Parking (AVP) \cite{becker2014bosch}.

The planning problem this work aims to solve can be stated as follows:
\par
\textit{"Find a solution in real-time that autonomously navigates a non-holonomic vehicle without any collisions, from a given start position to a desired goal position within the parking layout based on the input of a two-dimensional obstacle map, or report the non-existence of such a solution."}

To fully define the problem, the environment (obstacles), the vehicle model, and Key Performance Indicators (KPIs) must be defined.

\subsection{Environment}
\label{ch:ParkingLot}
%\begin{figure}
%	\centering
%	\includegraphics[width=\columnwidth]{figures/Config_Parallel.png}
%	\caption{\small Parallel Parking}
%	\label{fig:ParallelParking}
%\end{figure}

\begin{figure}
	\centering
	\includegraphics[width=\columnwidth]{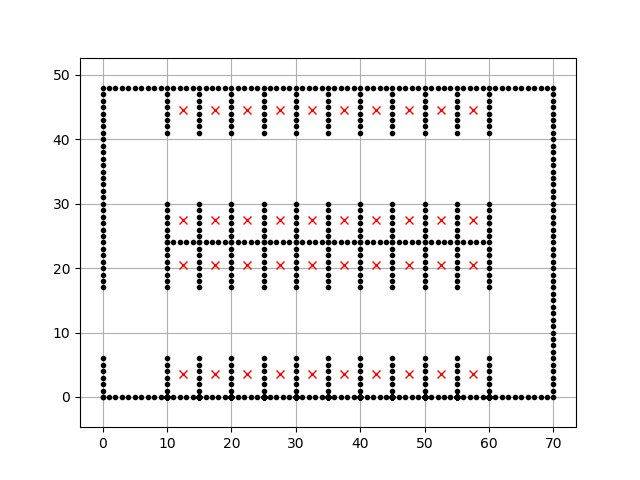}
	\caption{\small Parking with perpendicular parking slots}
	\label{fig:PerpendicularParking}
\end{figure}

Figure~\ref{fig:PerpendicularParking}, shows an example of the parking layout structure which is considered for the motion planning problem of autonomous parking. There might be two most prominent parking orientations, perpendicular or parallel. However, the algorithm should be general to work on different parking arrangements. Each parking slot is enumerated with a parking slot ID. The number of parking slots and dimensions of the open space are configurable. The red cross symbols in the figure highlight the goal position for each parking slot within the layout. 
The entry point to the parking lot is fixed at $(x_s, y_s,{\theta}_s) = (0, 10, 0)$. The parking position for the respective parking slots ID is read from a pre-computed map.
The computation of goal position within each parking slot is based on the vehicle dimensions as the control reference would be different for the individual vehicle due to differences in length and width.

\subsection{Vehicle Model}
\label{ch:Vehicle_Model}
Due to vehicle geometry and physics, there are constraints on the vehicle motion that restrict the allowable velocities. The first-order constraints, that consider the first derivative of the position (velocity), are often called \textit{kinematic} constraints. Including the dynamics of a vehicle results in second-order differential constraints, which allows the modeling of acceleration. The planning with such models is called \textit{kinodynamic} planning. As the focus of this work is on low-velocity parking maneuvers, higher-order constraints are not included, only kinematic constraints are considered.

%\begin{figure}
%	\includegraphics[width=\columnwidth]{figures/single_track_model.png}
%	\centering
%	\caption{\small Geometry of the basic single track or bicycle model of Car-like Robot}
%		\todo[inline]{Redraw}
%	\label{fig:Bicycle_Model}
%\end{figure}

%-----------------------------------------------------------------
\begin{figure}
	\centering
    \begin{subfigure}[t]{0.48\linewidth}
% \begin{figure}
        \centering
    	\includegraphics[width= \columnwidth]{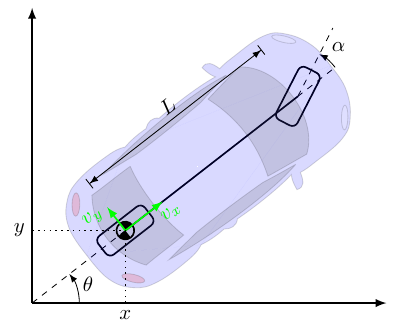}
        \subcaption{\small Single-track vehicle model.}
        \label{fig:Bicycle_Model}
    \end{subfigure}    
    \begin{subfigure}[t]{0.38\linewidth}
        \centering
    	\includegraphics[width= \columnwidth]{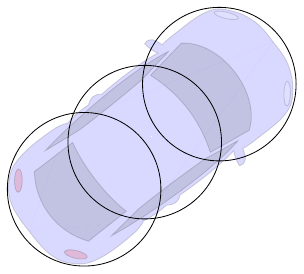}
        \subcaption{\small 3 Disk approximation for vehicle geometry.}
        \label{fig:Bouding_Volume_Used}
    \end{subfigure}
    \caption{Vehicle models.}
%-----------------------------------------------------------------
\end{figure}
%-----------------------------------------------------------------

A simple yet useful model of a car is the \textit{single track model}, also known as the \textit{bicycle model}, shown in Figure~\ref{fig:Bicycle_Model}. It does not consider  dynamics, but it is useful to model lower-velocity driving.
Consider the case where a car with wheelbase $L$ moves forward with velocity $v_{x}$ with a  steering angle $\alpha$ and assuming no wheel slip, then the car will move along a circle with radius $R$. The kinematic constraints can then be derived by trigonometry. 
Let $\textbf{x}$ = $(x,y, \theta)$ denote the configuration state of the car-like robot,  where $x$ and $y$ denote the position and $\theta$ the heading of the car. Vehicle motion is constrained to $\dot{x}/\dot{y} = \tan(\theta)$, which together with the constraint $R = {L}/{\tan(\alpha)}$ gives the following first-order differential constraints:
\begin{equation}
\dot{x} = v_{x}\cos(\theta)
\end{equation}
\begin{equation}
\dot{y} = v_{x}\sin(\theta)
\end{equation}
\begin{equation}
\dot{\theta} = \frac{v_{x}}{L}\tan(\alpha)
\end{equation}

Setting the maximum steering angle $|\alpha | \le {\alpha}_\mathrm{max}$ results in a minimum turning radius $R_\mathrm{min}$. The model clearly represents the non-holonomic behavior as it is impossible to move sideways without violating the no-slip condition. Restricting the allowed velocities and steering angles to the finite set $\mathcal{U}_{v_{x}} = \{0, 1\}$ and $\mathcal{U}_{\alpha} = \{-{\alpha}_\mathrm{max}, 0,{\alpha}_\mathrm{max}\}$ results in the Dubin’s car that can only stop and move forward at unit speed \cite{dubins1957on} and setting it to $\mathcal{U}_{v_{x}} = \{-1, 0, 1\}$ results in the Reed-Shepp car that can also reverse at a unit speed \cite{reeds1990optimal}. Even though these models are very simplified they have efficient analytic solutions for optimal paths between any two states that can be useful for designing heuristic functions for search-based motion planners.
%\cite{lavalle2006planning}

A time-discretized single-track model is easily obtained using Euler forward or higher-order methods. By integrating the differential equations forward in time, a simulated path or trajectory resulting from a given input can be obtained and used to construct motion segments within a planning framework.

%\begin{table}
%	\centering
%	\begin{tabular}{|l|l|}
%		\hline
%		Front / rear projection (mm) & 861/1.060 \\ \hline
%		Wheelbase (mm)               & 2.630     \\ \hline
%		Length (mm)                  & 4.551     \\ \hline
%		Width (mm)                   & 1.816     \\ \hline
%		Height at kerb weight (mm)   & 1.481     \\ \hline
%		Turning radius (m)           & 10.9      \\ \hline
%	\end{tabular}
%	\caption{\small SEAT LEON ST X-PERIENCE Vehicle Dimensions}
%	\label{Tab:VehicleDimensions}
%\end{table}

%For this work, in order to parameterize the vehicle model, the vehicle parameters of the production version of \textit{SEAT LEON ST X-PERIENCE} hatback car have been used which are provided in table~\ref{Tab:VehicleDimensions} and depicted in figure~\ref{fig:Vehicle_Configuration}.

%\begin{figure}
%	\includegraphics[width=\columnwidth, angle=0]{figures/Vehicle_Dimensions.png}
%	\centering
%	\caption{\small Reference Vehicle Configuration}
%		\todo[inline]{Redraw}
%	\label{fig:Vehicle_Configuration}
%\end{figure}

\subsection{Collision Detection}
\label{ch:MyCollDet}

%from intro
To avoid collisions with obstacles in the environment, vehicle geometry should be considered. The geometry of a car, approximate rectangular shape, can be reasonably approximated with overlapping circular disks. This simplifies the collision detection significantly as it is sufficient to check if the obstacles fall within the boundaries of the disk, which is represented by the radii of the disks. The problem of covering rectangles  with equal-sized disks in an optimal way has been studied in the literature \cite{melissen2000covering}.

As stated in \cite{ziegler2010fast}, a rectangle of length $l$ and width $w$ can be covered by $n$ circles of radius $r$ calculated as:
\begin{equation}
{r = \sqrt{\frac{l^2}{n^2} + \frac{w^2}{4}}}
\end{equation}

placed at a distance of $d$ calculated as:
\begin{equation}
{d = 2\sqrt{{r^2} - \frac{w^2}{4}}}
\end{equation}

In practical applications, the above approximation may lead to under-utilization of available collision-free space, especially in the case of environments where tight or narrow maneuvering is required, e.g. parking lots. Besides, the stated approach assumes the control reference to be at the center of the rectangular shape which is not true in the case of car-like robots which are mostly either front-wheel or rear-wheel driven.

In this work, an approach proposed in \cite{ziegler2010fast} is adapted to fit a practical application of a car-like robot. The bounding disks can be arranged as shown in Figure~\ref{fig:Bouding_Volume_Used}.
By this method, the bounding disks are arranged compactly to fit the geometry of the vehicle allowing better utilization of the free space even for tight maneuvers in narrow spaces. Moreover, all the calculations are based on the standard dimensions available from any production car design, and parametrizing the same makes the approach generic for any vehicle under consideration.

\subsection{KPI Definition for Benchmark}
\label{KPI}

To compare the performance and quality of solutions generated by motion planning algorithms, the following Key Performance Indicators (KPIs) have been used.

\textbf{Performance Paramters}

\textit{Number of Expanded States}: For a given configuration space, the number of expanded states reflects the guidance power of the heuristic functions in pruning the unwanted branches of the search. The lesser the number of expanded states, the better the heuristic.
	
\textit{Execution Time}: The execution time depends on the implementation of the vehicle model, the definition of motion primitives, and the algorithm itself. It is a measure of the time that the algorithm needs to return a solution using the defined attribute functions.

\textit{Number of Iterations}: The iteration counter quantifies how quickly the algorithm converges to either finding a solution or reporting that there exists no solution.

\textbf{Solution Path Quality Parameters}

\textit{Path Length}: The path length is computed as the accumulated sum of Euclidean distance between two points on the final trajectory. It quantifies the efficiency of the generated solution as shorter path lengths are preferred.

\textit{Reverse Path Length}: The reverse path length indicates the quality of the algorithm to foresee a wrong branch. The longer reverse path length indicates that the vehicle had to move a lot in a backward direction in order to correct its path or in some cases the algorithm prefers to move the vehicle more in a backward direction rather than forward. In any case, longer reverse path lengths are not preferred.

\textit{Direction Changes}: Each direction change during driving indicates a stop-and-go situation, which will be annoying for a human driver. Even though it is an autonomous vehicle the quality solution shall be close to an experienced human driver, i.e., avoid multiple direction changes.

\section{Motion Planning Approach}

The motion planning approach presented in this paper is based on Multi-Heuristic A* search, extended with a concept from the Hybrid A* Algorithm, employed in a bi-directional search fashion. The kinodynamic feasibility of the solution is provided by motion primitives based on the vehicle model. Several admissible heuristic functions enable efficient optimal planning.

\subsection{Hybrid A* Algorithm} %finished
\label{ch:Hybrid_Astar}

The Hybrid A* algorithm was developed as a practical path-planning algorithm that can generate smooth paths for an autonomous vehicle operating in an unstructured environment and used in the DARPA Urban Challenge by the Stanford University team \cite{dolgov2008practical}. The hybrid A* algorithm is based on the A* algorithm, with the key difference being, that state transitions occur in continuous rather than in a discrete space. By considering the non-holonomic constraints of the robotic vehicle, the algorithm generates feasible transitions which can be executed by the actuator module. 

The three-dimensional state space $\mathcal{X}$ (represented by $x,y$ position and $\theta$ heading angle of the vehicle) is associated with a discrete grid of reasonable resolution such that each continuous state is associated with some grid cell to enable the use of discrete search algorithm. Continuous states are rounded to the grid for association in order to prune the branches, by keeping only the best trajectory coming to the grid cell. The expansion still uses the actual continuous value that is not rounded to the grid. Similar to the original A*, if the current state being expanded is not the goal state, new successors are generated for all possible actions $u \in \mathcal{U}(\textbf{x})$. The \textit{cost-to-come} is only computed for successor states that are not in the $\textsc{Closed}$ list. If the state is not in the $\textsc{Open}$ list it is directly pushed to the $\textsc{Open}$ list. If the state is already in the $\textsc{Open}$ list, and the \textit{cost-to-come} is smaller than the cost for a state with the same index that is in the $\textsc{Open}$ list then the pointer to the parent, the \textit{cost-to-come} and the \textit{cost-to-go} are updated. After that, the key is decreased using the newly computed cost.

\subsection{Multi Heuristic A* Algorithm} %finihsed
\label{ch:MHA*}

The performance of the A* algorithm depends on the quality of the heuristic function used to guide the search. It is hard to design a single heuristic function that captures all the complexities of the problem. Furthermore, it is hard to ensure that heuristics are admissible (provide lower bounds on the cost-to-go) and consistent, which is necessary for an A*-like search to provide guarantees on completeness and bounds on sub-optimality.

%maximum of all heuristics

In \cite{roger2010more} authors introduced an approach of alternation between different heuristic functions for satisficing (i.e. non-optimal) planning. Multi-Heuristic A* (MHA*) \cite{aine2016multi} overcomes the dependency on a single heuristic function in optimal planning too. MHA* can use multiple inadmissible heuristic functions in addition to a single consistent heuristic simultaneously to search in a way that preserves guarantees on completeness and bounds on sub-optimality. This enables us to effectively combine the guiding powers of different heuristic functions and simplifies dramatically the process of designing heuristic functions by a user because these functions no longer need to be admissible or consistent \cite{aine2016multi}.

MHA* has two variants: Independent Multi-Heuristic A* (IMHA*) which uses independent cost-to-come and cost-to-go values for each search, and Shared Multi-Heuristic A* (SMHA*) which uses different cost-to-go values but a single cost-to-come value for all the searches. With this shared approach, SMHA* can guarantee the sub-optimality bounds with at most two expansions per state. In addition, SMHA* is potentially more powerful than IMHA* in avoiding depression regions as it can use a combination of partial paths found by different searches to reach the goal \cite{aine2016multi}.

In SMHA* approach the optimal path for a given state is shared among all the searches so that if a better path to a state is discovered by any of the searches, the information is updated in all the priority queues. This allows the algorithm to expand each state at most twice, which significantly improves the computational time.

%\todo[inline]{ALGORITHM}
\subsection{Our Planning Algorithm}
\label{ch:Framework}

\begin{algorithm}[t]
	\scriptsize
	\SetAlgoVlined
	%	\SetAlFnt{\small}
	%	\SetAlCapFnt{\small}
	%	\SetAlCapNameFnt{\small}
	\SetKwInOut{Input}{input}\SetKwInOut{Output}{output}
	\SetKwFunction{FSMHSearch}{$\texttt{SMHA}^\ast$}%
	\SetKwFunction{FCombPath}{$\texttt{CombinePath}$}%
	\SetKwProg{Fn}{function}{:}{}%
	
	%\Input{$x_{start}$, $x_{goal}$, $\mathcal{O}$}
	%\Output{Collision free trajectory from $x_{start}$ to $x_{goal}$}
	
	\Fn{\FSMHSearch{$x_\mathrm{I}$, $X_\mathrm{G}$, $\mathcal{O}$,$ h_{i}$}}
		{
			\KwRet{Collision free trajectory from $x_\mathrm{I}$ to $x \in X_\mathrm{G}$}
		}
	
%	\Fn{\FGenPath{$x_{start}$, $x_{goal}$, $\mathcal{O}, h_{i}$, ${d_\mathrm{fw}}$}}
%		{
%			\While{$\mathit{EuclideanDistance(x, x_{goal})}$ $\geq$ ${d_\mathrm{fw}}$}
%			{
%				$\mathit{SharedMultiHeuristicA^\ast(x_{start}, x_{goal}, \mathcal{O}, h_{i})}$
%			}
%			\KwRet{Collision free trajectory from $x_{start}$ to $x_{goal}$, $x_{current}$}
%		}
	
	\Fn{\FCombPath{$(x_\mathrm{start},\dots, x_\mathrm{f}')$,$(x_\mathrm{f}'', \dots, x_\mathrm{G})$}}
		{
			\tcc{Use an analytic function to connect paths}
			\KwRet{Combined trajectory from $x_\mathrm{start}$ to $x_\mathrm{goal}$}
		}
	
	\Begin
	{
		\tcc{Forward Search}
		$X_\mathrm{G}\leftarrow \{x \; | \; \lVert x-x_\mathrm{G} \rVert \leq d_\mathrm{fw1} \} $\;
		$(x_\mathrm{start},\dots, x_\mathrm{f}')$ $\leftarrow$ $\texttt{SMHA}^\ast(x_\mathrm{start}, X_\mathrm{G}, \mathcal{O}, h_{i})$\;
		\tcc{Reassign Start and Goal Positions}
		$x_\mathrm{start}'$ $\leftarrow$ $x_\mathrm{goal}$\;
		$X_\mathrm{G}'\leftarrow \{x \; | \; \lVert x-x_\mathrm{f}' \rVert \leq d_\mathrm{fw2} \} $\;
		\tcc{Backward Search}
		$(x_\mathrm{G}, \dots, x_\mathrm{f}'')$ $\leftarrow$ $\texttt{SMHA}^\ast(x_\mathrm{start}', X_{G}',
		\mathcal{O}, h_{i})$\;
		\tcc{Combine Paths}
		$path \leftarrow \texttt{CombinePath}((x_\mathrm{start},\dots, x_\mathrm{f}'),(x_\mathrm{f}'', \dots, x_\mathrm{G}))$\;
	}	
	\KwRet{path}
	
	\caption{\small Bi-Directional Multi-Heuristic Search}
	\label{Alg:SMHAstar_Algorithm}
\end{algorithm}

The presented planning algorithm is based on  MHA* Search with features of Hybrid A* search and adaptive motion primitives.  It uses MHA* in a forward and backward manner as shown in Algorithm \ref{Alg:SMHAstar_Algorithm}. For this purpose, multiple admissible and non-admissible heuristic functions are defined.
 %
%The framework enables the verification of the developed method to targeted use-cases and benchmark benefits of developed planners compared to the other state-of-the-art approaches. The framework has a generic structure to allow easy reuse, modifications or extension of the algorithm based on future requirements. 
%For benchmarking, comparisons to related state-of-the-art approaches are performed mainly based on Key Performance Indicators (KPIs) defined in section~\ref{KPI}.
%
The framework has three main functions \texttt{SharedMultiHeuristicA*}, \texttt{GeneratePath}, and  \texttt{CombinePath}.
\texttt{SharedMultiHeuristicA*} searches the configuration space according to the algorithm defined in \cite{aine2016multi}. It is expanded with hybrid A* features and uses motion primitives. It is used in both forward and backward steps.
\texttt{GeneratePath} executes the search using \texttt{SharedMultiHeuristicA*} while continuously checking if the \texttt{EucledianDistance} from the current state to the goal state is greater than a configurable parameter $d_\mathrm{fw}$. If the search has reached the closest state defined by $d_\mathrm{fw}$ then the function returns the path from the start position to the closest point to the goal.
Due to the discretization of the continuous space, in \texttt{CombinePath} an analytic function is required to combine the two paths generated by \texttt{GeneratePath} function. In this work, we used Reeds-Shepp curves for this purpose.

\subsubsection{Bi-Directional Search} %finished
The presented algorithm is using the \texttt{Bi-directional Search}, by searching for the path in two steps. In the first step, the search expands in the forward direction (towards the goal) from the start state to reach the state close to the goal position. In the second step, the search proceeds backward from the goal position toward the closest point reached by the forward search. The solution paths generated by the forward search step and backward search step are then joined by the analytical expansion using Reeds-Sheep curves.

%Bidirectional
%Multiheuristic
%Adaptive size motion primitives

\subsubsection{Motion Primitives} %finished
\label{MotionPrimitives}
The motion primitives refer to the motion sequence that is triggered by an action request and corresponds to a basic move that is possible by the vehicle, sampled from a continuous control space. In this work, motion primitives are generated by applying one of the six control actions defined by combinations of $\mathcal{U}_{v_{x}} = \{-1, 1\}$ and $\mathcal{U}_{\alpha} = \{-{\alpha}_\mathrm{max}, 0,{\alpha}_\mathrm{max}\}$. These represent maximum steering left while driving in the forward direction, no steering while driving in the forward direction, maximum steering right while driving in the forward direction, maximum steering left while driving in the backward direction, no steering while driving in the backward direction, maximum steering right while driving in the backward direction. Finer resolution is also possible, however that increases the branching factor and computational complexity.

%\begin{figure}
%	\includegraphics[width=\columnwidth, angle=0]{figures/Motion_Primitives.png}
%	\centering
%	\caption{\small Motion Primitives Used for Thesis}
%		\todo[inline]{Redraw}
%	\label{fig:Motion_Primitives}
%\end{figure}

%-----------------------------------------------------------------
\begin{figure}
\centering
	\includegraphics[width= 0.7 \columnwidth]{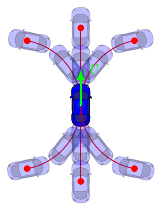}
  \caption{\small Motion Primitives.}
  \label{fig:Motion_Primitives}
\end{figure}
%-----------------------------------------------------------------

As shown in Figure~\ref{fig:Motion_Primitives}, each of these control actions is applied for a certain amount of time, resulting in an arc of a circle with a lower bound turning radius $R_\mathrm{min}$. This will ensure that the resulting paths are always drivable, as the actual vehicle model is used to expand the state, even though they might result in excessive steering actions. 
An \textit{\textbf{adaptive sizing of motion primitives}} is applied, wherein the arc length used for the execution of motion primitives is adapted dynamically to adjust to the environment. A shorter arc length is used near obstacles and a longer arc length in free space. This approach improves maneuverability in tight spaces. Using a shorter length in all cases promises higher levels of resolution completeness, as the likelihood to reach each state is increasing but reduces the computational efficiency.

\subsubsection{Heuristics} %finished
\label{ch:Heuristics}
A heuristic function $\mathit{h}$ is used to estimate the cost needed to travel from some state $x$ to the goal state $x_{g}$ (cost-to-go). As it is shown in \cite{hart1968formal}, if the heuristic function is underestimating the optimal cost-to-go, A* search provides the optimal solution. For the shortest path search, the usual heuristic function is the Euclidean distance.

In general, SMHA* algorithm supports $n$ number of heuristics with $n > 1$. In this work, to restrict the complexity and to be comparable with Hybrid A* which is used as a reference for benchmarking, two heuristic functions have been used. The two heuristics capture different aspects of the problem as explained in the sections below.

\begin{figure}
	\centering
    \begin{subfigure}[t]{0.48\linewidth}
        \centering    	
    	\includegraphics[width= \columnwidth]{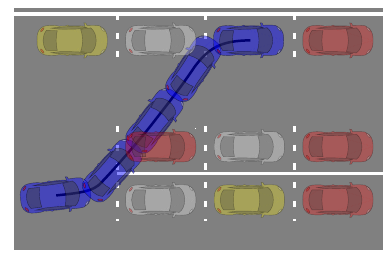}
    	\subcaption{\small Non-holonomic without obstacles (Reeds-Shepp Curve).}
    	\label{fig:heuristics2}
    \end{subfigure}
% \end{figure}
% \begin{figure}    
    \begin{subfigure}[t]{0.48\linewidth}
    	\centering
    	\includegraphics[width= \columnwidth]{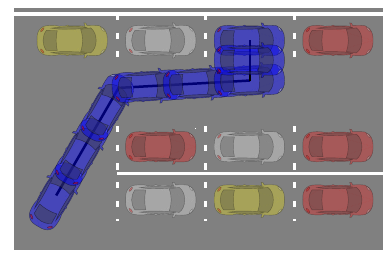}
    	\subcaption{\small Holonomic with obstacles.}
    	\label{fig:heuristics1}
    \end{subfigure}
    \caption{Paths based on relaxed models of heuristic functions.}
\end{figure}

\textbf{Non-Holonomic without Obstacles} %finished
This heuristic function takes into account the non-holonomic constraints of the vehicle while neglecting the influence of the environment (obstacles). The most suitable candidate functions are either Dubins or Reeds-Shepp curves. These curves are the paths of minimal length with an upper bound curvature for the forward and combined forward and backward driving car respectively. We choose the Reeds-Shepp curves since in parking maneuvers it is important that the car can move in both forward and backward directions. These curves are computationally inexpensive to compute as they are based on an analytic solution.
%\begin{figure}
%	\includegraphics[width=\columnwidth, angle=0]{figures/Figure_Reeds_Shepp_1.png}
%	\centering
%	\caption{\small Reeds Shepp Curve (Steering Left, Right, Left)}
%	\label{fig:Figure_Reeds_Shepp_1}
%\end{figure}
As shown in Figure~\ref{fig:heuristics2}, this heuristic takes into account the current heading as well as the turning radius, which ensures that the vehicle approaches the goal with the appropriate heading. This is especially important when the car gets closer to the goal. Given that Reeds-Shepp curves are minimal, this heuristic is clearly admissible.

\textbf{Holonomic with Obstacles} %finished
This heuristic function neglects the characteristics of the vehicle and only accounts for obstacles. The estimate is based on the shortest distance between
the goal state and the state currently being expanded. This distance is determined using the standard Dijkstra search in two dimensions ($x$ and $y$ position). As the search is 2D and assumes the object under control is holonomic, the path is not smooth. The search is performed backward, it uses the initial state of the SMHA* as the goal state, and the goal state of the SMHA* search as the start state to generate the heuristic cost. The closed list of the Dijkstra search stores all the shortest distances to the goal and guides the vehicle away from dead ends and around obstacles. Since this heuristic function does not depend on any runtime sensor information, it can be fully pre-computed offline and used as a lookup table or simply translated and rotated to match the current goal instead of initiating a new search while SMHA* progresses.

%\begin{figure}
%	\centering
%		\centering
%		\includegraphics[width= \columnwidth]{figures/DJ_Figure_1_Costmap.png}
%		\caption{\small Example Cost Map of Dijkstra Search}
%		\label{fig:DJ_Figure_1_Costmap}
%\end{figure}
%\begin{figure}
%		\centering
%		\includegraphics[width= \columnwidth]{figures/DJ_Figure_2_ShortPath.png}
%		\caption{\small Example Path Generated - Dijkstra Search}
%		\label{fig:DJ_Figure_2_ShortPath}
%	\caption{\small Dijkstra Search}
%	\label{fig:Dijkstra Search}
%\end{figure}

%Figure~\ref{fig:DJ_Figure_1_Costmap} shows an example cost map that depicts the cost heat map of the cost-to-go to reach the goal state. Figure~\ref{fig:DJ_Figure_2_ShortPath} depicts the shortest path from a given start to the goal position generated by the Dijkstra search. 

\section{Simulation results}

In order to benchmark the performance of the solution developed using SMHA*, we chose the Hybrid A* as a reference. As discussed earlier, Hybrid A* gives a comparable reference as it is also based on the orderly sampling approach and also uses two heuristics to guide the search.
The key difference is that in the Hybrid A* approach, the maximum of both of the results of the heuristics is considered to update the priority queue while in SMHA* the heuristics are iteratively computed and both can update the priority queue.

\label{ch:software}
The use cases chosen for the simulation depict common situations encountered in a parking lot such as \textit{Entering Parking Lot} and \textit{Exiting Parking Lot}. The simulations are performed by executing the Hybrid A* and SMHA* algorithm back-to-back to compare the KPIs of the generated solution.

\begin{figure}[t!]
    \centering
    \begin{subfigure}[t]{0.49\linewidth}
		\centering
    	\includegraphics[width=0.9\columnwidth]
    	{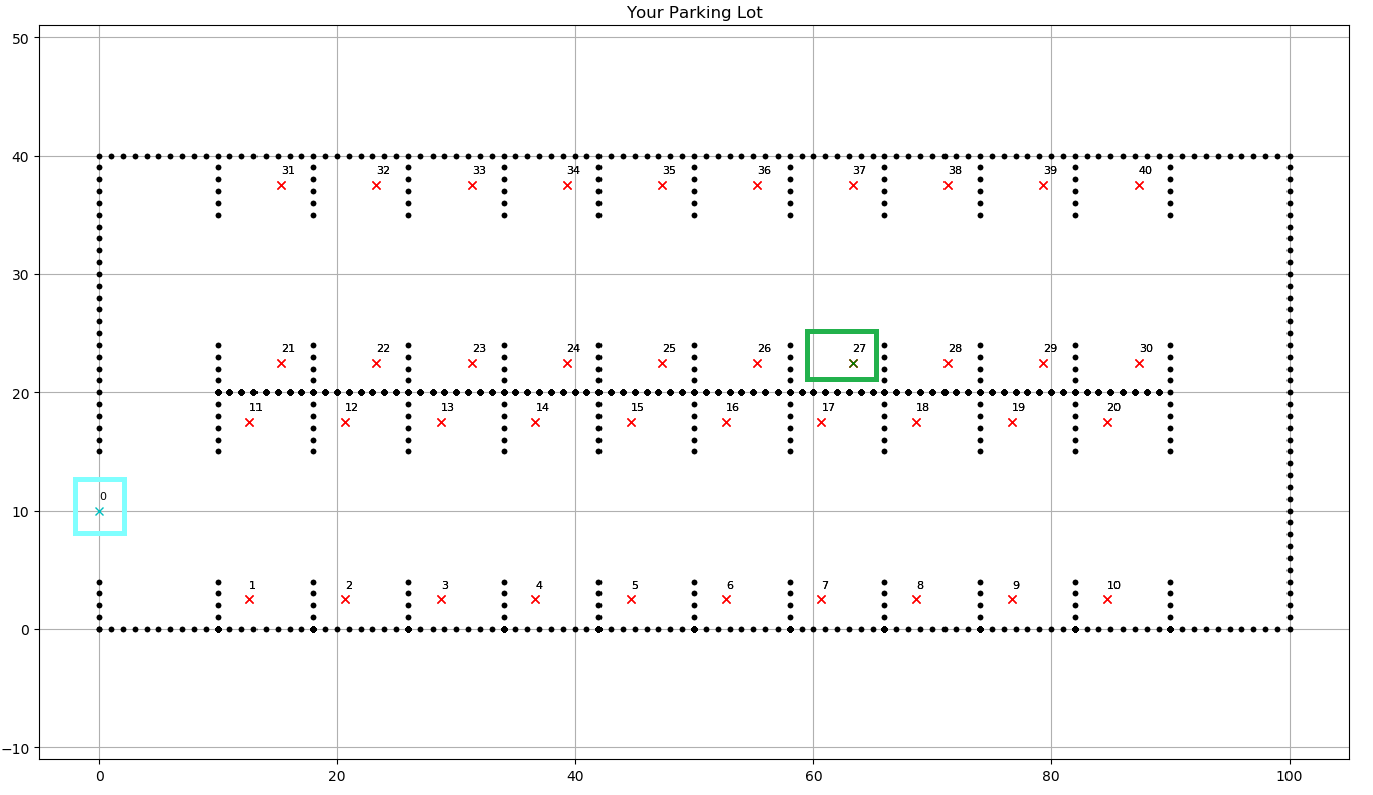}
    	\centering
    	\subcaption{\small .}
    	\label{fig:UC1_BWD_Figure1_ParkingSlot_0_27}
    \end{subfigure}
    \begin{subfigure}[t]{0.49\linewidth}
    	\includegraphics[width=0.9\columnwidth]
    	{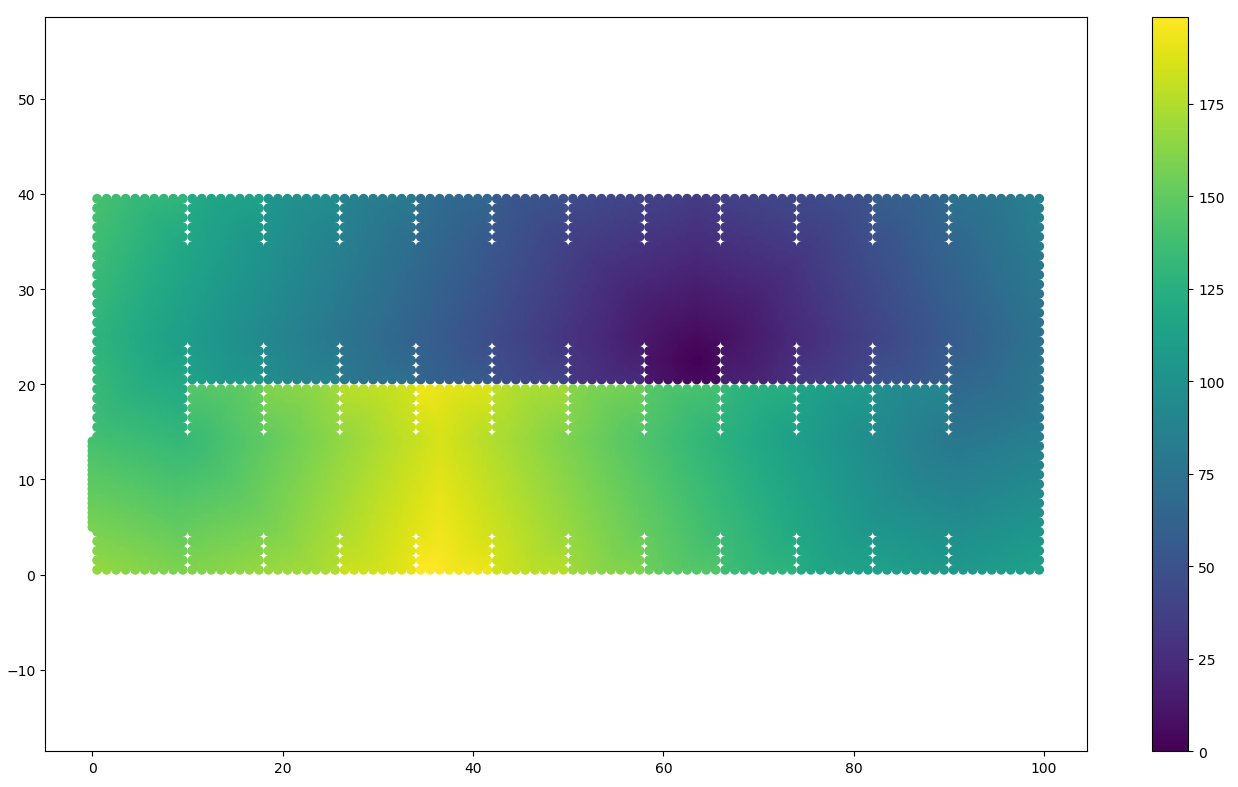}
    	\centering
    	\subcaption{\small }
    	\label{fig:UC1_BWD_Figure2_ParkingSlot_0_27}
    \end{subfigure}
    \caption{Parking Lot with Start Position - Cyan and Goal Position - Green (left) and cost-to-go map generated by 2D Dijkstra search (right).}
\end{figure}

In the rest of the section, an elaborate analysis of the simulation results and solution paths that are generated for the use case \textit{Entering Parking Lot} using  \texttt{Bi-directional Search} for parallel parking configuration is presented. Figure~\ref{fig:UC1_BWD_Figure1_ParkingSlot_0_27} depicts the selected start and goal position (Parking Slot ID: 27) on the parking layout.

First, the configuration space is explored using the 2D Dijkstra search to generate the cost-to-go map as shown in Figure~\ref{fig:UC1_BWD_Figure2_ParkingSlot_0_27}. The cost-to-go map shows in a color scale distance from the goal pose considering obstacles but neglecting non-holonomic constraints. The results of this step are stored in a look-up and represents the \textit{Holonomic with Obstacles} heuristic explained earlier.

\begin{figure}[t!]
    \centering
    \begin{subfigure}[t]{0.49\linewidth}
		\centering
		\includegraphics[width=0.9\columnwidth]
		{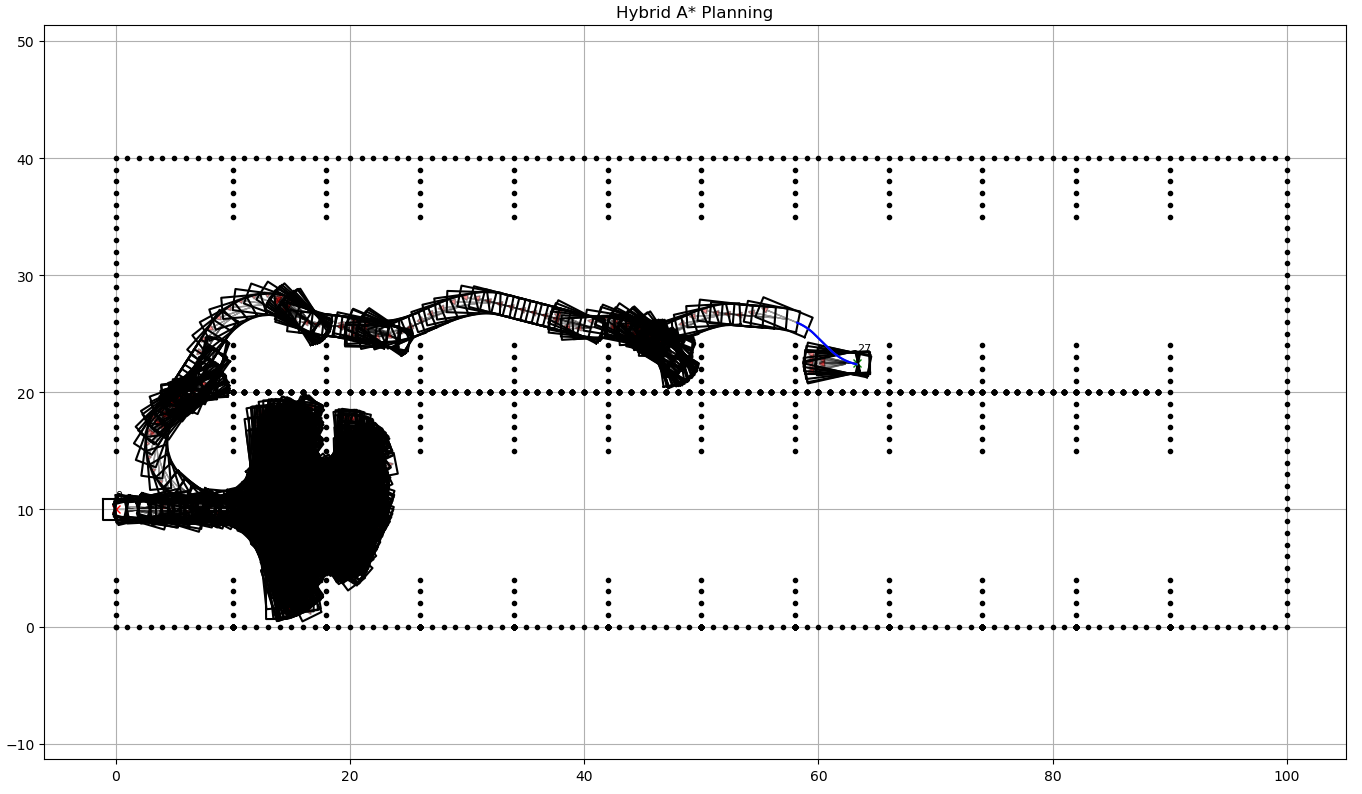}
		\subcaption{\small State expansions.}
		\label{fig:UC1_BWD_Figure3_ParkingSlot_0_27}
    \end{subfigure}
    \begin{subfigure}[t]{0.49\linewidth}
		\centering
		\includegraphics[width=0.9\columnwidth]
		{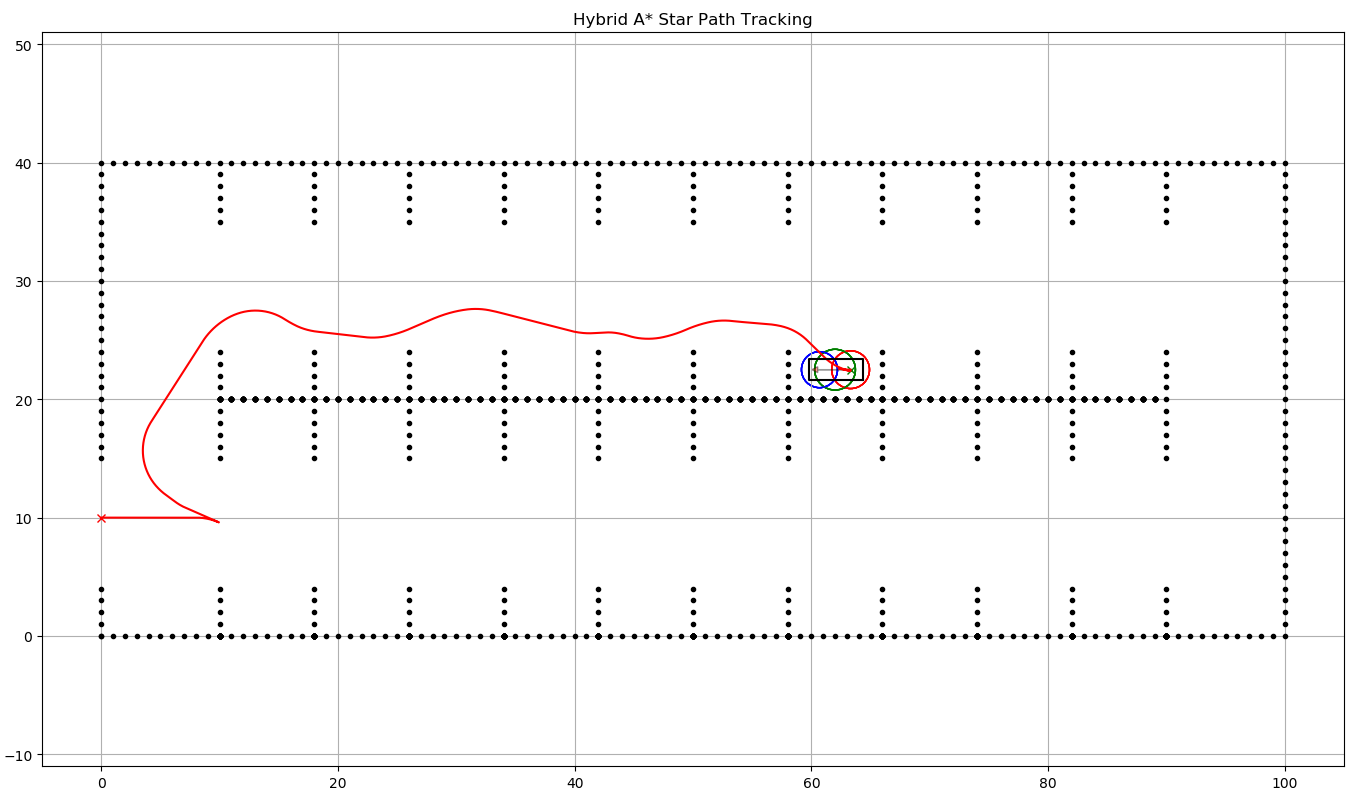}
		\caption{\small Path generated by Hybrid A*.}
		\label{fig:UC1_BWD_Figure4_ParkingSlot_0_27}
    \end{subfigure}
	\caption{\small Entering Parking Lot with Bi-Directional - Hybrid A*.}
	\label{fig:Entering Parking Lot BWD Hybrid A*}
\end{figure}

Figure~\ref{fig:UC1_BWD_Figure3_ParkingSlot_0_27} depicts the state expansion pattern of the Hybrid A* algorithm. The algorithm has searched the area around the start position with a bias towards the goal position even though the solution path lies in another direction. The heuristic strongly guides the search towards the shortest path as far as possible within the obstacle-free area. As the search explores, it expands all states with lower costs that can lead to the shortest path until it reaches a point where the heuristic cost of expanding the points which do not lead to the shortest path has a lower cost to reach the goal. As a result of this poor pruning of the unwanted branches, the planner expands several states around the start position before it realizes the optimal direction of the path which leads to poor timing performance.
The final path generated as seen in Figure~\ref{fig:UC1_BWD_Figure4_ParkingSlot_0_27} has many orientation changes in the \texttt{Forward Search} step and maneuvering step into the parking slot is determined by the backward search. The solution path is smooth, but the direction of orientation is reversed for the most part of the path which is not optimal.

\begin{figure}[t!]
    \begin{subfigure}[t]{0.5\linewidth}
% \begin{figure}
		\centering
		\includegraphics[width=0.9\columnwidth]
		{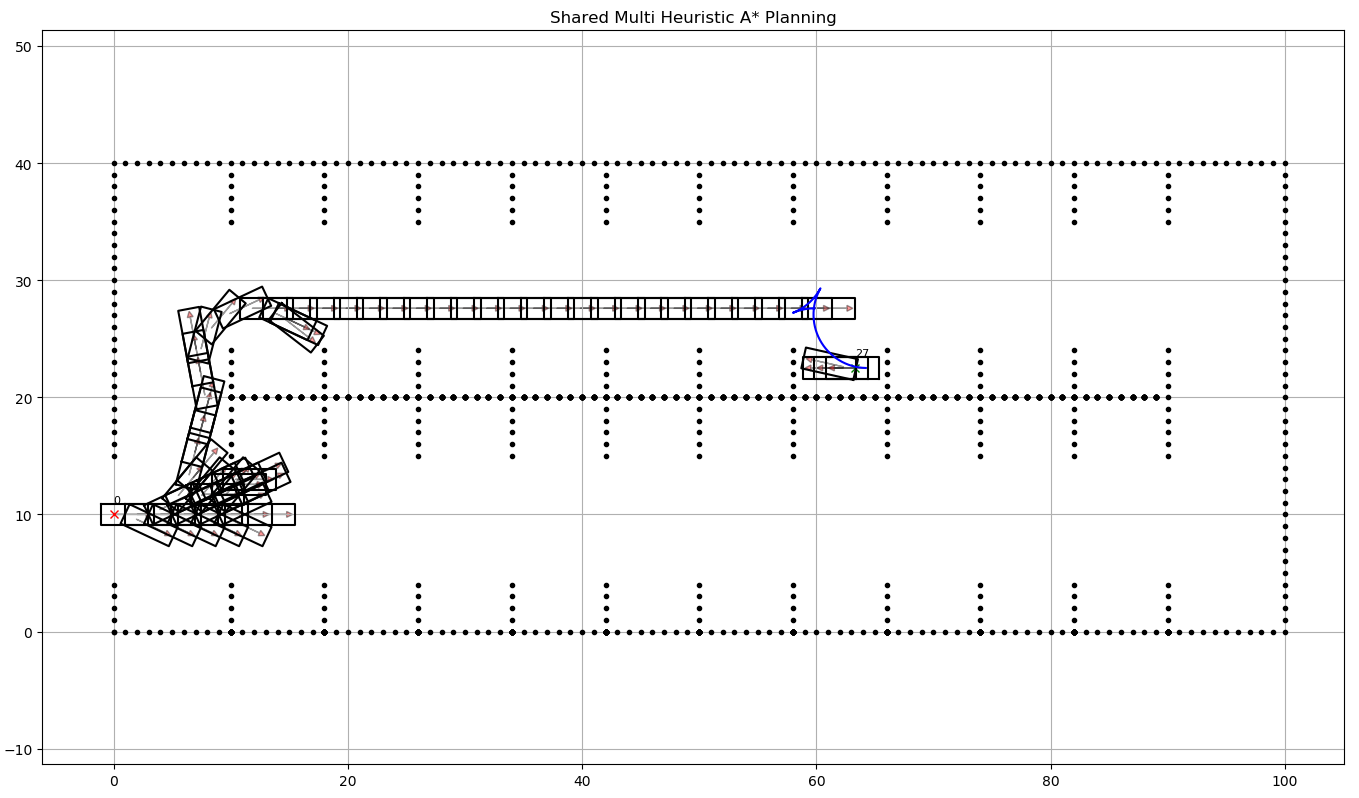}
		\caption{\small State expansions.}
		\label{fig:UC1_BWD_Figure5_ParkingSlot_0_27}
    \end{subfigure}%
% \end{figure}
	\begin{subfigure}[t]{0.5\linewidth}
% \begin{figure}
		\centering
		\includegraphics[width=0.9\columnwidth]
		{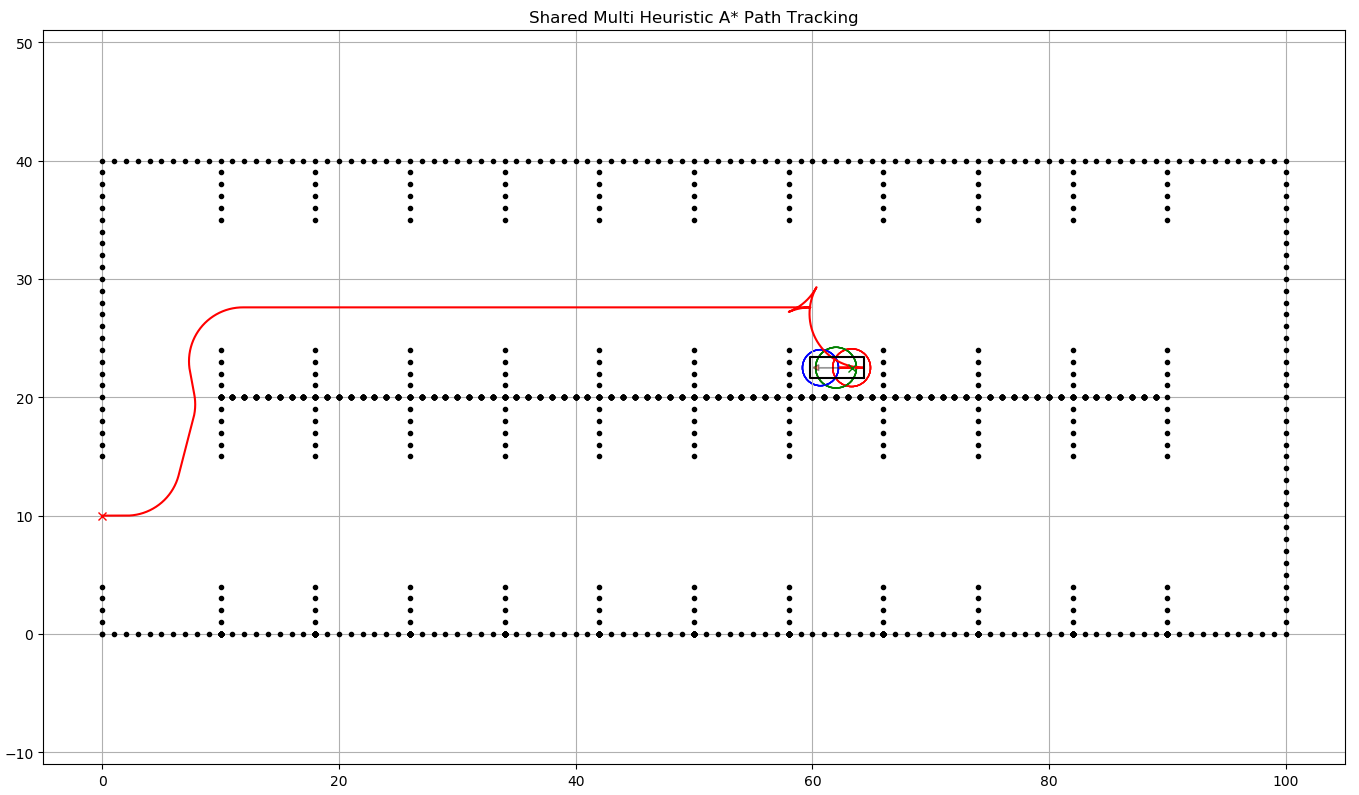}
		\caption{\small Path generated.}
		\label{fig:UC1_BWD_Figure6_ParkingSlot_0_27}
    \end{subfigure}
	\caption{\small Entering Parking Lot with Bi-Directional - SMHA*.}
	\label{fig:Entering Parking Lot BWD SMHA*}
% \end{figure}
\end{figure}

Figure~\ref{fig:UC1_BWD_Figure5_ParkingSlot_0_27} depicts the state expansion pattern of the SMHA* algorithm. Similar to Hybrid A*, the algorithm has searched the area around the start position with a bias towards the goal position even though the solution path lies in another direction. But, the multi-heuristic approach, quickly balances the bias towards finding the shortest path to finding a feasible path considering the obstacles. In addition, due to the mutually informed independent search by respective heuristics, the states that are expanded by one heuristic function are not expanded by other heuristic functions. As a result, the algorithm could prune the unwanted branches and realize the optimal direction of the path much faster compared to Hybrid A*.

As seen in Figure~\ref{fig:UC1_BWD_Figure6_ParkingSlot_0_27}, the path is smooth and the direction of orientation is in the forward direction for most parts of the path, which is preferred.

\begin{table}[t]
	\scriptsize
	\centering
	\begin{tabular}{|c|c|c|}
		\hline
		%		\rowcolor[HTML]{FFFFC7} 
		\textbf{KPI} & \textbf{Hybrid A*} & \textbf{SMHA*} \\ \hline
		        Number of Expanded States          &        2457        &       73        \\ \hline
		            Execution Time (s)             &        47.8        &      11.51      \\ \hline
		             Path Length (m)               &        90.3        &      90.58      \\ \hline
		         Reverse Path Length (m)           &        7.29        &      3.29       \\ \hline
		            Direction Changes              &         4          &        4        \\ \hline
		           Number of Iterations            &       23633        &       73        \\ \hline
	\end{tabular}
	\caption{\small Entering parking lot with Bi-Directional Search - KPI Comparision.}
	\label{Tab:Entering Parking Lot BWD KPI Comparision}
\end{table}

\begin{figure}[h]
	\includegraphics[width=0.9\columnwidth]                   
	{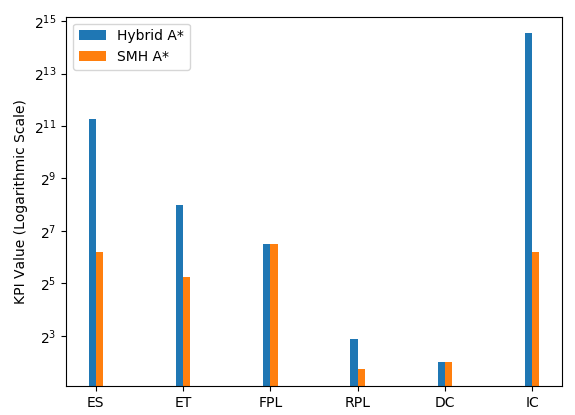}
	\centering
	\caption{\small Entering Parking Lot with Bi-Directional Search - KPI Comparision (Expanded States (ES), Execution Time (ET), Forward Path Length (FPL), Reverse Path Length (RPL), 	Direction Changes (DC), Iteration Count (IC)).} 
	\label{fig:UC1_BWD_Figure20_ParkingSlot_0_27}
\end{figure}

As seen from KPI values tabulated in Table~\ref{Tab:Entering Parking Lot BWD KPI Comparision}, the Hybrid A* algorithm expands significantly more states and uses more time to generate the solution path compared to SMHA* approach. Even though the heuristics used are the same, the mutually informed independent search of SMHA* prunes the unwanted branches  much more significantly allowing faster convergence towards the solution and improved execution time. 

To give a comprehensive performance comparison of both the algorithms for the use case \textit{Entering Parking Lot} for parallel parking lot layout using \texttt{Bi-directional Search}, a full simulation run through all the parking slots is executed. In this simulation mode, each parking slot ID is selected as a goal position sequentially and the back-to-back run of the Hybrid A* and SMHA* algorithm is performed.

\begin{figure}[h]
	\includegraphics[width=0.9\columnwidth]
	{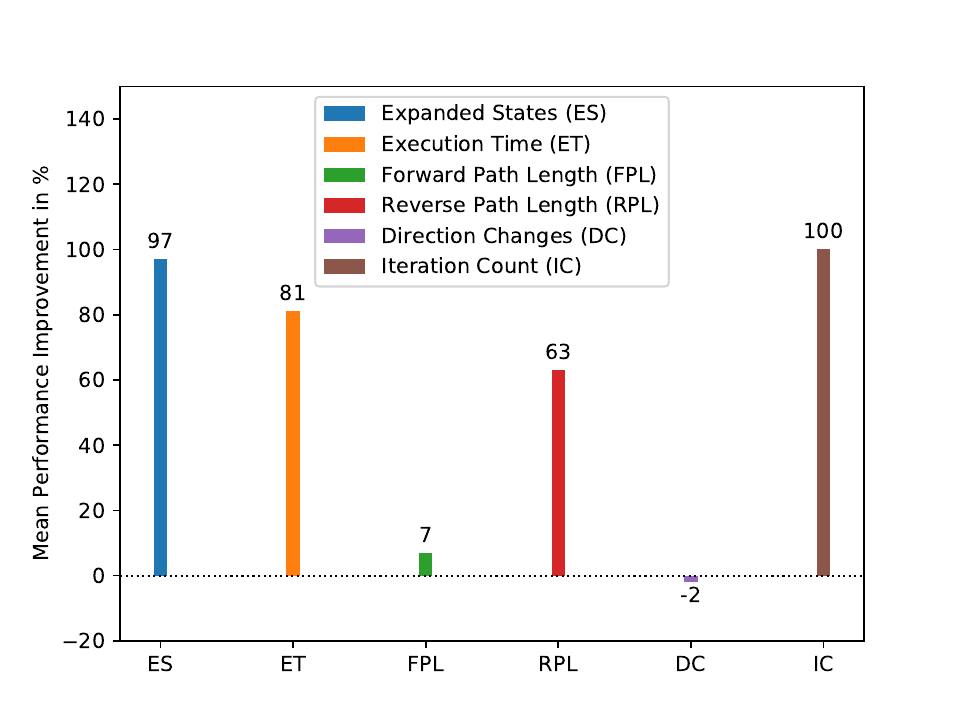}
	\centering
	\caption{\small Entering Parking Lot with Bi-Directional Search - Mean Performance Improvement.} 
	\label{fig:Enter_BWD_Peformance_Summary_0}
\end{figure}

As observed from Figure~\ref{fig:Enter_BWD_Peformance_Summary_0}, SMHA* outperforms the Hybrid A* algorithm in terms of both performance and solution path quality parameters. With the multi-heuristic approach,  the average execution time to generate the solution path is reduced by \textbf{81\%}, which is a significant improvement demonstrating the potential of multi heuristic approach to solve the given planning problem.	
	
\section{Conclusion} %finished
The work focused on providing a Multi-Heuristic search-based approach to solve the motion planning problem for autonomous parking. To benchmark the results obtained, a state-of-the-art planning algorithm Hybrid A* was chosen as a reference.

As the environment for the given use case involves only low-speed maneuvering, a Single Track Bicycle Model which reflects the kinematics of the vehicle was used as a motion model. The model emulates the non-holonomic nature of the vehicle in all stages of the algorithm, motion primitives (node expansion), and heuristic estimates. Thus, the paths generated are always driveable.

A collision check algorithm was implemented based on the Multi-Disk Decomposition of the bounded volume. The algorithm was parameterized based on the vehicle geometry making it a generic solution to fit with any vehicle type and independent of the environment.

The path planning problem was solved using the Shared Multi-Heuristic A* approach in which two heuristic functions were implemented with a round-robin scheduling. The respective heuristic searches share the current path obtained to a state. The heuristics were defined to capture the non-holonomic and holonomic constraints of the vehicle. Two solution approaches were developed: \texttt{Forward Search} and \textit{Bi-Directional Search }.

The SMHA* algorithm solved the motion planning problem elegantly and outperformed Hybrid A* with respect to the response time and path quality of the generated solution path. The KPI comparison clearly indicates that SMHA* is an ideal candidate for motion planning in slow-speed driving in unstructured environments applications like autonomous valet parking.

\section{Acknowledgments}

The project leading to this study has received funding from the European Union’s Horizon 2020 research and innovation programme under the Marie Skłodowska-Curie grant agreement No 675999, ITEAM project.\par
VIRTUAL VEHICLE Research Center is funded within the COMET – Competence Centers for Excellent Technologies – programme by the Austrian Federal Ministry for Transport, Innovation and Technology (BMVIT), the Federal Ministry of Science, Research and Economy (BMWFW), the Austrian Research Promotion Agency (FFG), the province of Styria and the Styrian Business Promotion Agency (SFG). The COMET programme is administrated by FFG.\par

\bibliographystyle{IEEEtran} % use IEEEtran.bst style
\bibliography{bibfile}

\end{document}